\documentclass[conference]{IEEEtran}
\IEEEoverridecommandlockouts
\usepackage{cite}
\usepackage{amsmath,amssymb,amsfonts,nccmath}
\usepackage{graphicx}
\usepackage{textcomp}
\usepackage{xcolor}

\usepackage{bm}
\usepackage[noend]{algpseudocode}
\usepackage{algorithmicx}
\usepackage{algorithm}
\usepackage{multirow}
\usepackage{booktabs}
\usepackage{pifont}
\usepackage{verbatim}
\usepackage{balance}
\usepackage{footmisc}
\usepackage{amssymb}
\usepackage{graphics}

\def\BibTeX{{\rm B\kern-.05em{\sc i\kern-.025em b}\kern-.08em
    T\kern-.1667em\lower.7ex\hbox{E}\kern-.125emX}}
\begin{document}

\title{CSAFL: A Clustered Semi-Asynchronous Federated Learning Framework
}

\author{\IEEEauthorblockN{Yu Zhang\IEEEauthorrefmark{1}, 
		Moming Duan\IEEEauthorrefmark{1},
		Duo Liu\IEEEauthorrefmark{1},
		Li Li\IEEEauthorrefmark{1},
		Ao Ren\IEEEauthorrefmark{1},
		Xianzhang Chen\IEEEauthorrefmark{1},
		Yujuan Tan\IEEEauthorrefmark{1},
		Chengliang Wang\IEEEauthorrefmark{1}
	}
	\IEEEauthorblockA{\IEEEauthorrefmark{1}College of Computer Science, Chongqing University, Chongqing, China\\
	}
	\IEEEauthorblockA{Email: zhangyucqu9@gmail.com,	 $\{$duanmoming, liuduo$\}$@cqu.edu.cn
		}

}

\maketitle

\setlength\footnotemargin{0em}
\let\thefootnote\relax\footnotetext{ \rule[0.25\baselineskip]{0.5\columnwidth}{0.5pt}\\
	This paper will be presented at IJCNN 2021.}

\begin{abstract}
Federated learning (FL) is an emerging distributed machine learning paradigm that protects privacy and tackles the problem of isolated data islands. 
At present, there are two main communication strategies of FL: synchronous FL and asynchronous FL. The advantages of synchronous FL are that the model has high precision and fast convergence speed. However, this synchronous communication strategy has the risk that the central server waits too long for the devices, namely, the straggler effect which has a negative impact on some time-critical applications. Asynchronous FL has a natural advantage in mitigating the straggler effect, but there are threats of model quality degradation and server crash. Therefore, we combine the advantages of these two strategies to propose a clustered semi-asynchronous federated learning (CSAFL) framework. We evaluate CSAFL based on four imbalanced federated datasets in a non-IID setting and compare CSAFL to the baseline methods. The experimental results show that CSAFL significantly improves test accuracy by more than $+5\%$ on the four datasets compared to \textit{TA-FedAvg}. In particular, CSAFL improves absolute test accuracy by $+34.4\%$ on non-IID FEMNIST compared to \textit{TA-FedAvg}.
\end{abstract}


\section{Introduction}

Federated learning (FL)~\cite{mcmahan2017communication,zhao2018federated,li2020federated} is an emerging machine learning framework that utilizes multiple edge devices to jointly train a global model under the coordination of the central server. 
The training process of FL is divided into plenty of communication rounds. 
In each communication round, the edge devices download and use the global model parameters from the central server to perform optimization with local data for obtaining the local model parameters. 
Finally, each edge device updates the local model parameters to the central server for aggregating new global model parameters. 
Throughout the training process of the federated network, the training data of the client is always kept locally and is not transmitted, which protects data privacy~\cite{privacy2019,privacy2020,privacy2021}. 
In practice, The FL framework plays a crucial role in supporting privacy-sensitive applications on edge devices~\cite{privacyapp}.

\begin{figure}[t]
	\centering
	
	\includegraphics[width=8.9cm]{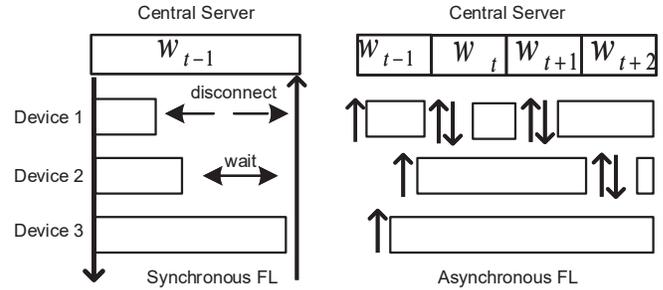}
	\vspace{-8mm}
	\caption{The training procedures of synchronous FL and asynchronous FL}
	\label{fig:flframework}
\end{figure}

There are two model updating mechanisms for federated learning: synchronous and asynchronous. As shown in Fig.~\ref{fig:flframework}, we observe the training process of the two update strategies.  
Due to the different update mechanisms, these two FL frameworks have their own merits and drawbacks:

\begin{itemize}
	\item Synchronous FL: In the synchronous FL, all clients need to download the global model parameters at a unified time node, and the central server waits for all clients to complete the training tasks. 
	The merits of this synchronous updating strategy are the model has fast convergence and high precision. 
	The only drawback is the straggler effect~\cite{stragglereffect} caused by network resources or poor hardware resources, which puts the server fall into an idle state.
	\item Asynchronous FL: Under the asynchronous FL, the central server cooperates with clients that complete the training tasks, rather than waiting for all clients, and each client can ask the central server for the new global model parameters when the training procedure is completed. Although this model updating strategy avoids the server falling into the idle state, it also has some shortcomings. On the one hand, this strategy makes the data transmission larger, which may cause the server to crash~\cite{device}. On the other hand, the gradient divergence caused by asynchronous updating will further degrade the performance of the model.
\end{itemize}

There have been many efforts in FL synchronous or asynchronous training algorithms. 
McMahan \textit{et al.} propose the FL framework \textit{Federated Averaging (FedAvg)}~\cite{mcmahan2017communication} and Li \textit{et al.} experimentally and theoretically prove that \textit{FedAvg} can achieve $\mathcal{O}(1/T)$ convergence rate with decayed learning rate under a statistical heterogeneous setting~\cite{Li2020On}. To relieve the effect of imbalanced data on accuracy, Duan \textit{et al.} propose a self-balancing FL framework Astraea~\cite{duan2019astraea}. \textit{FedAvg} and Astraea are both synchronous FL frameworks. However, these algorithms do not consider the stragglers, which is unfriendly for time-critical applications. 
Asynchronous FL framework mitigate the straggler effect. 
Xie \textit{et al.} Proposed \textit{FedAsync}~\cite{xie2019fedasyn} framework with a staleness function. Chen \textit{et al.} Propose an asynchronous algorithm to tackle the challenges associated with heterogeneous edge devices, such as computational loads, stragglers~\cite{aso-fed}. However, these frameworks ignore the situation that the server may crash due to the continuous processing of requests from the clients~\cite{device}.

In this paper, we propose a novel clustered semi-asynchronous federated learning (CSAFL) framework, which mitigates the straggler effect and controls model staleness in the asynchronous update for accuracy improvement. 
CSAFL leverages a spectral clustering algorithm~\cite{spectral}, which groups clients according to the affinity matrix constructed by the delay and direction of clients’ model update. In each communication round with fixed time budget, the selected clients do synchronous or asynchronous update independently, and contribute the update parameters to the group model them belong to. 
The main contributions of this paper are summarized as follows:
\begin{itemize}
	\item We propose a novel clustered semi-asynchronous federated learning (CSAFL) framework. As far as I know, this is the first framework that combines synchronous and asynchronous update mechanisms.
	\item We design two strategies to alleviate the model staleness problem caused by asynchronous updating. The first is to leverage a spectral clustering algorithm to divide clients with different learning objectives into multiple groups. The second is by limiting the model delay.
	\item We evaluate CSAFL on four federated datasets, and show its effectiveness in mitigating the straggler effect compared to \textit{T-Fedavg}, and show the accuracy improvement of the model is more than $+5\%$ on the four datasets compares to \textit{TA-FedAvg}. Specially, the maximum accuracy improvement is $+34.4\%$ on FFEMNIST dataset compared to \textit{TA-FedAvg}.
\end{itemize}

The rest of this paper is organized as follows.
Section~\ref{sec:background} outlines the background of synchronous FL and asynchronous FL.
Section~\ref{sec:motivation} shows the motivation of the CSAFL framework. Section~\ref{sec:design} details the design of the CSAFL framework. Evaluation results are presented and analyzed in Section~\ref{sec:evaluation}. 
Section~\ref{sec:conclusion} concludes the paper.

\section{background and related work}\label{sec:background}

\subsection{Federated Learning}
In this section, we first present the vanilla synchronous FL framework \textit{FedAvg}. Then we briefly introduce the asynchronous FL framework based on Fedavg.

McMahan \textit{et al.} first propose the concept of federated learning, and later propose a widely used federated learning framework \textit{FedAvg}~\cite{mcmahan2017communication}, which involves solving a machine learning problem by loosely combining multiple devices under the coordination of a central server. 
Unlike the traditional distributed machine learning method, the computing nodes in the FL framework keep the training data locally and do not exchange or transfer the training data in the federated network. 
Instead, only the local updates of each computing node are transmitted, which reduces privacy risks. 
Nevertheless, the central server has no control over computing nodes. For example, computing nodes can join or drop out of the federated network at any time. 
\begin{algorithm}[t]
	\caption{Federated Averaging}
	\label{alg:fedavg}
	\begin{algorithmic}[1]
		\Procedure {FL Server Training}{}
		\State Initialize global model $\bm{w}_0$.
		\For{each communication round $t=1,2,...,T$}
		\State $S_t \leftarrow$ Randomly select $K$ clients from $N$ clients.
		\State Server broadcasts $\bm{w}_t$ to the selected clients. 
		\For{each client $c_i \in S_t$ parallelly}
		\State $\bm{w}_{t+1}^i\leftarrow$~\textbf{ClientUpdate($i$,~$\bm{w}_t$)}.
		\EndFor
		\State $\bm{w}_{t+1}\leftarrow\bm{w}_t+\sum_{c_i \in S_t} \frac{n_i}{n} \bm{w}_{t+1}^i$
		\EndFor
		\EndProcedure
		\Statex
		\Function{ClientUpdate}{$i$,~$\bm{w}$}
		\State $\hat{\bm{w}}\leftarrow\bm{w}$
		
		\For{each local epoch $e=1,2,...,E$}
		\For{each local batch $b \in \mathcal{B}$}
		\State $\bm{w}\leftarrow\bm{w}-\eta\nabla L(b;\bm{w})$
		\EndFor
		\EndFor
		\State \Return $\bm{w}$
		\EndFunction
	\end{algorithmic}
\end{algorithm}
In particular, the optimization goal of general FL is:

\begin{equation}
	\label{fml:fw}
	\min\limits_{\bm{w}}f(\bm{w})\triangleq \sum_{k=1}^{N}p_k F_k(\bm{w})
,
\end{equation}
which means solving model updates $\bm{w}$ when the value of $f(\bm{w})$ is the smallest. In~\eqref{fml:fw}, $N$ is the number of clients participating in training, $p_k$ refers to the participation weight of the $k$-th device, and $p_k \geqslant 0$, $\sum_{k}^{} p_k = 1$, $F_k(\bm{w})$ is the local objective optimization function of the $k$-th device, we define $F_k(\bm{w})$ as:

\begin{equation}
	F_k(\bm{w})\triangleq \frac{1}{n_k}\sum_{(x_i,y_i) \sim p_{data}^k}l(x_i,y_i,\bm{w})
,
\end{equation}
where $n_k$ is the training data size of the $k$-th client, $(x_i,y_i)$ is the sample of the $k$-th client and obeys the data distribution $p_{data}$. $l(x_i,y_i,\bm{w})$ is the prediction loss function on $(x_i,y_i)$.

Specifically, the synchronous FL based on \textit{FedAvg} mainly includes a central server maintaining the global model and multiple participants, which communicate through network connections with the server. In each communication round $t$, the server first selects a part of the client $K$ from all clients $N$ to participate in the training tasks and then broadcasts the global model $\bm{w_t}$ through the network into the selected clients. The client $c_i$ updates the $\bm{w_{t+1}^i}$ locally based on $\bm{w_t}$. When the client completes the local training procedure, it transmits $\bm{w_{t+1}^i}$ to the server through the network. The server waits for all clients to complete the training task. Finally, the server aggregates the new global model $\bm{w_{t+1}}$ by averaging the local solutions of the clients. For an FL task, to achieve the target accuracy, it usually needs hundreds of communication rounds.

The pseudo-code of $FedAvg$ is shown in Algorithm~\ref{alg:fedavg}, where $S_t$ is a random subset of $K$ clients randomly selected in each communication round. Minibatch $\mathcal{B}$ is the batch number of training data, and $b$ refers to the subset of training data separated by $\mathcal{B}$. Local epoch $E$ is the number of local training. $n_i$ is the data size of client $c_i$ and $n=\sum_{c_i \in S_t} n_i$ is the total data size of selected clients. $\eta$ is the learning rate of the local solver. 

\renewcommand{\algorithmicrequire}{\textbf{Input:}}  
\renewcommand{\algorithmicensure}{\textbf{Output:}} 

As shown in Algorithm~\ref{alg:fedavg}, we note that the server needs to wait for all clients to complete the synchronous FL system's training tasks, which is negative for some time-critical applications. 
Therefore, the asynchronous FL framework is proposed. We briefly introduce the asynchronous FL framework based on \textit{FedAvg}, where each client updates the local model to the global model independently. Whenever the server receives the local update from the client, it will refresh the global model. Therefore, the server does not need to wait for stragglers for aggregation.


\subsection{Related Work}
Federated Learning (FL), first proposed by Google, is a new approach to fitting machine learning into the edge. 
Existing FL frameworks can be classified into synchronous FL and asynchronous FL according to model updating. 
Most of the studies on synchronous FL do not consider the issues of stragglers due to the device heterogeneity and the instability of network conditions~\cite{mcmahan2017communication}~\cite{noconsider2}~\cite{noconsider3}. 
To eliminate the straggler effect on the statistical heterogeneity, Li \textit{et al.}~\cite{li2018federated} propose a near term, experimentally prove this term can improve the stability of the framework and provide convergence guarantees in theory. However, the server still needs to wait for the local updates of stragglers before aggregation. 
In order to address the issue that all clients have to wait for the slowest one, Li \textit{et al.}~\cite{lifedavgnoniid} propose a strategy to let the central server only accept the responses of the top threshold k clients, and the rest of the clients are regarded as stragglers. This method simply dropouts the stragglers, ignores the possibility of valuable data on the stragglers. 
In addition, to mitigate the impact of stragglers, Dhakal \textit{et al.}~\cite{codefl} develop a  coded computing technique for FL where sever compensates the gradient of the stragglers. However, this compensation is the result of the sever calculates the gradients based on the client's parity data. There is a risk of privacy leakage. 

Asynchronous FL has a natural advantage over synchronous FL in solving the straggler effect~\cite{safafed}~\cite{fedasyn20201}~\cite{fedasyn2020}, where the server can aggregate without waiting for stragglers. 
Chen \textit{et al.} propose a \textit{ASO-Fed} framework~\cite{aso-fed}, which updates the global model in an asynchronous manner to tackle stragglers. However, gradient staleness is not considered, which may threaten the convergence of the model. 
Especially, Xie \textit{et al.} develop a \textit{FedAsync}~\cite{xie2019fedasyn} algorithm which combines a function of staleness with asynchronous update protocol. However, the clients will continue to transmit a large amount of data to the server, which may cause the server to crash. 
In terms of reducing data transmission, Wu \textit{et al.} propose a \textit{SAFA}~\cite{safafed} protocol which divides clients into three types, in which asynchronous clients continuously perform local update until the difference between local update version and global model version reaches tolerance. 
Although \textit{SAFA} considers model staleness, the server needs to wait for the asynchronous clients.

\section{preliminary case studies and motivation}\label{sec:motivation}
In this section, we illustrate our preliminary case studies, which guide the motivation for a new grouping model update strategy in FL. 

\begin{figure}[b]
	\centering
	\vspace{-5mm}
	\includegraphics[width=8.9cm]{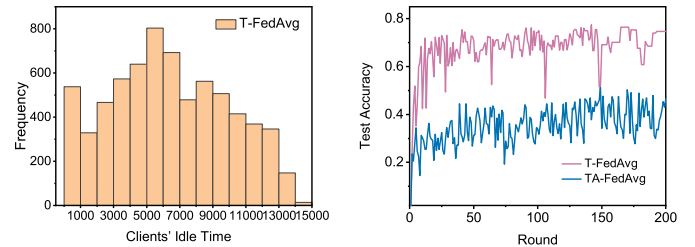}
	\vspace{-5mm}
	\caption{The straggler effect in the synchronous FL framework can be observed in the figure on the left, and the frequency distribution of the clients‘ idle time is unbalanced; The figure on the right shows the accuracy comparison of asynchronous FL and synchronous FL in the process of model training.}
	\label{fig:gmotivation}
\end{figure}

Asynchronous FL has a natural advantage in solving the straggler effect in FL~\cite{safafed}, but the model delay caused by asynchronous update may affect the accuracy of the model. To investigate the influence of different updating strategies on the accuracy of the model influence of different model updating strategies, we study the training process of synchronous FL and asynchronous FL. More specifically, we set up this test based on a 62- class Federated Extended MNIST~\cite{caldas2018leaf} (FEMNIST) dataset using the multinomial logistic regression model. The statistical information of FEMNIST dataset is shown in Table~\ref{table1}. In order to show the straggler effect clearly, we set a fixed time budget hyper-parameter for each communication round. So in a communication round, the client may do multiple synchronous or asynchronous updates. Under a fixed time budget, we set up two groups of experiments based on \textit{FedAvg}, which are \textit{T-FedAvg} and \textit{TA-FedAvg}. \textit{TA-FedAvg} is an asynchronous update algorithm based on \textit{FedAvg}. In this FL test, we adopt the same notation as~\cite{mcmahan2017communication}: the number of all clients K = 200, the size of local minibatch $\mathcal{B} = 10$, the number of local epochs $E = 10$, the number of round $T = 200$, the learning rate $\eta = 0.03$, the number of selected clients per round is $20$. We add the time budget of each round $H$. In this test, $H = 15000ms$.

\begin{table}[t]
	\centering  
	\caption{The statistics of the four federated datasets}  
	\label{table1}  
	\begin{tabular}{cclcc}
		\hline
		Task                                  & Dataset            & \multicolumn{1}{c}{Model} & devices & samples \\ \hline
		\multirow{2}{*}{Image Classification} & MNIST              & \multirow{3}{*}{MCLR}     & 1000    & 69035   \\ \cline{2-2} \cline{4-5} 
		& FEMNIST            &                           & 200     & 18345   \\ \cline{1-2} \cline{4-5} 
		Simulation Test                       & Synthetic(0.8,0.5) &                           & 100     & 75349   \\ \hline
		Sentiment Analysis                    & Sent140            & LSTM                      & 772     & 40783   \\ \hline
	\end{tabular}
\end{table}
The top-$1$ test accuracy of the two model update strategies based on FEMNIST dataset is shown in Fig.~\ref{fig:gmotivation}. For the \textit{TA-FedAvg} algorithm, a $26.3\%$ reduction in the top-$1$ test accuracy compared to the \textit{T-FedAvg} algorithm. This test shows that the asynchronous update strategy harms the accuracy of the model. In the case of synchronous model update, all clients simultaneously download the global model parameters from the server. There is no client training with old model parameters in the same communication round. Obviously, the model delay of each client is $1$. However, in the case of asynchronous model update, When the server receives the local model parameters uploaded by one client, the global model on the server has been updated many times by other clients. Therefore, we realize that the model delay of each client is not steady. Due to the instability of model delay, the accuracy of asynchronous updates has decreased significantly.

As shown in Fig.~\ref{fig:gmotivation}, we can observe the clients' idle time distribution, which clearly shows that the server needs to wait too long, and the idle time is unbalanced in synchronization FL. However, in asynchronous FL, the server does not need to wait for all clients to complete the training tasks of each round. That is, when a client finishes uploading local model parameters, the server immediately refresh the global model, so the idle time of each client is $0$. We assume that the communication time for the client to download model parameters from the server is negligible. 

The preliminary case study shows that there is a straggler effect in the synchronous model update. In addition, the asynchronous model update strategy has the effect of precision degradation caused by model delay. We are also inspired by the Iterative Federated Clustering Algorithm (IFCA) framework proposed by Avishek~\cite{Avishek2020cluster}, which considers different groups of users have their objectives. Therefore, we propose a novel clustered semi-asynchronous federated learning (CSAFL) framework.

\section{CSAFL}\label{sec:design}
\subsection{Framework Overview}
To tackle the straggler effect in synchronous FL and model staleness in asynchronous FL, we propose a novel clustered semi-asynchronous federated learning (CSAFL) framework, which leverages a clustering algorithm based on similarity metrics to group clients. CSAFL combines synchronous FL and asynchronous FL to drive clients to update local model parameters to the same group's group model. 
As far as we know, this is the first paper to combine synchronous and asynchronous update mechanisms. 

\begin{figure}[t]
	\centering
	\includegraphics[width=8.9cm]{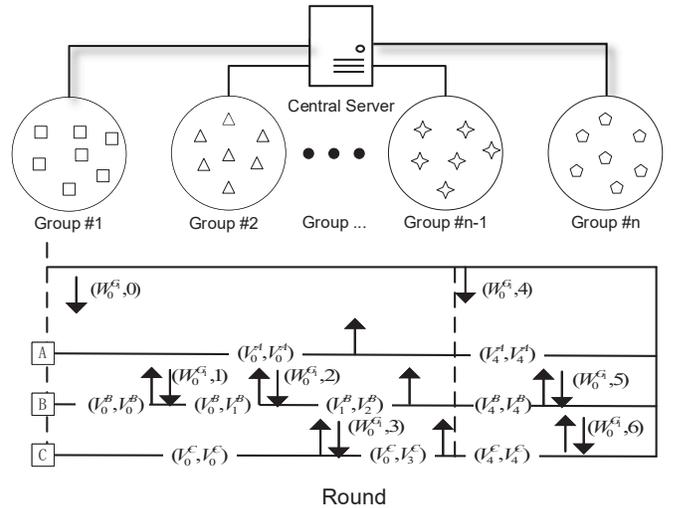}
	\vspace{-8mm}
	\caption{The framework of CSAFL. In a group training process, group 1 randomly selects three clients, A, B, C. The up arrow indicates that the client updates the local model to the group model, and the down arrow indicates that the client downloads the parameters of the group model}
	\label{fig:framework}
\end{figure}

Our model architecture is shown in Fig.~\ref{fig:framework}, CSAFL includes several groups, which can be deployed to the central server, or some devices in the middle layer, such as the edge server. 
They maintain the group models. 
In a group, the group model is the global model and the latest model. 
In this paper, we assume that all groups are deployed on a central server. 
CSAFL also contains a number of clients, which can be mobile phones, IoT devices, etc. 
There is a one-to-one communication between the client and its group. Due to the asynchronous update mechanism, the model maintained on the client is not necessarily the latest. 

The training process of CSAFL is shown in Algorithm \ref{alg:fedspecasync}, \textit{CSAFL} first initializes the group model $W_0^G$. 
In each communication round, the selected clients update the group model in its group, as shown in Fig.~\ref{fig:framework}. 
For group $g_x$, clients $A, B, C$ are the selected clients in one communication round. 
The update process within the group $g_x$ is divided into the following five steps: 
\begin{enumerate}
	\item Group $g_x$ broadcasts the initial model to clients $A, B, C$ and passes the version number of the group model, which is represented by ($W^{g_x}, 0$)
	\item After the clients $A, B, C$ receive the latest group model, they independently update the version number of the local model, such as, for client $A$ ($V_{pre}^A, V_{new}^A$)
	\item Clients $A, B, C$ asynchronously update the group model according to its computing capacities and communication conditions. 
	Each process of the asynchronous update will increase the version number of the group model. 
	Before update operation, the difference between $V_{pre}$ and $V_{new}$ is calculated. If the difference is more significant than the tolerance $H$ of gradient staleness, the clients whose time budgets are not exhausted are forced to update synchronously. 
	\item After each synchronization update, all clients download the same group model. {\label{item:4}}
	\item Repeat step $2$ to step $4$ until the time budget of the communication round is exhausted. 
\end{enumerate}

Similarly, when other clients are selected, the training process within the group is the same. 
Each group completes its own intra group update without interference. 
Each group has the same number of communication rounds. 

We cover more details of CSAFL in the following four sections. 

\subsection{Clustering Strategy}
\noindent$\displaystyle \textbf{Clustering 	Algorithm}$

As shown in Algorithm \ref{alg:fedspecasync}, we apply spectral clustering in this paper. Spectral clustering is more suitable for our framework, the following reasons:
\begin{itemize}
	\item It is more general than the K-Means clustering algorithm~\cite{kmeans}, concretely, it is suitable for both convex sets and non-convex sets. 
	\item It is not very sensitive to stragglers. 
	\item It uses the Laplacian Eigenmaps for high-dimensional data, which reduces the load of calculation.
\end{itemize}
\noindent$\displaystyle \textbf{Similarity Metrics}$

The traditional clustering federated learning (CFL) framework is based on the similarity of sub-optimization tasks, which only considers the similarity of gradient direction~\cite{cfl1}~\cite{cfl2}. But in FL, this is not comprehensive. For example, In the asynchronous FL, we assume that the network conditions are same, and the transmission capacities of the devices are also fixed, and the device $c_A$ with high computing power and the device $c_B$ with poor computing power are divided into a group, the $c_A$ will constantly refresh the global model with its own local updates, and the result of model training is more inclined to the $c_A$. In the synchronous FL, this grouping strategy will lead to the model training inefficient. Similarly, the data amount of different devices and different network conditions will make this grouping strategy have a bad impact on the performance of the model. Therefore, the data amount and the computing power of the device, and the network condition between the devices and the server may also be indicators of devices’ similarity. The above three indicators are represented by the local model update latency of the client, so we determine the metrics for measuring the clients' similarity: gradient direction and the latency of model update. In the local update of a client, the local update latency can be divided into two parts: computation latency and communication latency, and we define the following three delay models:

\subsubsection{Computation Latency}
To quantify the randomness of the client’s computing capabilities, we use a shift exponential distribution~\cite{shift}:
\begin{equation}\label{fml:exp}
	P[t_{cp}^{i}<t]=\left\{
	\begin{array}{rcl}
		1-e^{-\frac{\mu_{i}}{d_{i}}(t-x_{i} d_{i})} & & {t\geq x_{i} d_{i}}\\
		0 & & {otherwise}
	\end{array} \right.
,
\end{equation}
where $x_{i}>0$ and $\mu_{i}>0$ are parameters related to the computing capabilities of the client $c_{i}$. 
$x_{i}>0$ and $\mu_{i}>0$ represent the maximum and fluctuation of the client $c_{i}$ computation capabilities, respectively. 
$d_{i}$ is the size of the dataset on the client $c_{i}$, and $t$ is the computation latency of the client $c_{i}$. Because the physical facility's computing capabilities where the group is located is much greater than that of the client, We ignore the delay due to group model aggregation. 
Based on~\eqref{fml:exp}, we can get the mathematical expectation of computation latency for client $c_{i}$ as follows: 
\begin{equation}\label{fml:expcp}
	t_{excp}^i=x_{i} d_{i}+\frac{d_{i}}{\mu_{i}}	,
\end{equation}

\begin{algorithm}
	\caption{clustered semi-asynchronous federated learning}
	\label{alg:fedspecasync}
	\begin{algorithmic}[1]
		\Require
		number of all clients $R$, number of selected clients per round $K$, set of groups $G$, learning
		rate ${\eta}$, local minibatch size $\mathcal{B}$, number of local epochs $E$,global initial model 
		$\bm{w}_{0}$, time budget  of each communication round $H$, number of round $T$, delay threshold per round $L$.
		\Ensure
		updated group model parameters $\mathcal{W}_{T}^G$.		
		\Procedure {FedSpecAsync Training}{}
		\State Initialize group model.
		\State for all $g_i{\in} G$, initial $\bm{w}_{0}^{g_{i}}$ to $\bm{w}_{0}$
		\State $\mathcal{W}_{0}^G\leftarrow[\bm{w}_{0}^{g_{1}},...,\bm{w}_{0}^{g_{n}}]$	.	
		\For{each communication round $t=1,2,...,T$}
		\State $S_t \leftarrow$ Sever selects random $K$ clients from all clients $R$.
		\State for all $g_i{\in} G$, initial $v_{g_i}$ to $0$, $S_{t}^{g_{i}}\leftarrow\{c_j\mid c_j{\in}g_{i}.clients,\forall c_j{\in}S_{t}\}$.
		\For{each group ${g}_{i}$ in $G$ parallelly}
		\State for all $c_j{\in} S_{t}^{g_{i}}$, initial $v_{pre}^{c_j}$ to $0$, $v_{new}^{c_j}$ to $0$, $\bm{w}_{last}^{c_{j}}$ to $\bm{w}_{0}^{g_{i}}$
		\For{each client $c_j$ in $S_{t}^{g_{i}}$ parallelly}
		\While{$H$ is not exhausted}
		\State $\Delta{v}\leftarrow v_{new}^{c_j}-c_{j}^{prev}$.
		\If{$\Delta{v}>L$}
		\State $S_{syn}\leftarrow \{c_j\mid$ the $H$ of $c_j$ is not exhausted, $\forall {c_j} {\in}S_{t}^{g_{i}}\}$.
		\State ~\textbf{SynUpdates($S_{syn}$)}
		\Else
		\State $\bm{w}_{last}^{c_{j}}\leftarrow $~\textbf{ModelUpdates($c_{j}$,~$\bm{w}_{last}^{c_{j}}$)}.
		\State $\bm{w}_T^{g_i}\leftarrow \bm{w}_{last}^{c_{j}}$.
		\State ${v}_{g_i}\leftarrow {v}_{g_i}+1$.
		\State ${v}_{prev}^{c_j}\leftarrow {v}_{new}^{c_j}$.
		\State ${v}_{new}^{c_j}\leftarrow {v}_{g_i}$.
		\EndIf
		\EndWhile
		\EndFor
		\EndFor
		\EndFor
		\EndProcedure
		\Statex
		\Function{ModelUpdates}{$c$,~$\bm{w}$}
		\State $\mathcal{B}\leftarrow$ split dataset of client $c$ into batches of size $B$.
		\For{each local epoch $e$ from $1$ to $E$}
		\For{each local batch $b \in \mathcal{B}$}
		\State $\bm{w}\leftarrow\bm{w}-\eta\nabla L(b;\bm{w})$.
		\EndFor
		\EndFor
		\State \Return $\bm{w}$.
		\EndFunction
		\Statex
		\Function{SynUpdates}{$S_{syn}$}
		\For{each client $c_{z}$ in $S_{syn}$ parallelly}
		\State $\bm{w}_{last}^{c_{z}}\leftarrow $~\textbf{ModelUpdates($c_{z}$,~$\bm{w}_{last}^{c_{z}}$)}
		\State ${v}_{g_i}\leftarrow {v}_{g_i}+1$.
		\EndFor
		\State $n\leftarrow \sum_{c_z \in S_{syn}} {n^{c_z}}$.
		\State $\bm{w}_{last}^{g_{i}}\leftarrow \sum_{c_z \in S_{syn}}\frac{n_i}{n}\bm{w}_{last}^{c_{z}}$.
		\For{each client $c_{z}$ in $S_{syn}$ parallelly}
		\State $\bm{w}_{last}^{c_{z}}\leftarrow \bm{w}_{last}^{g_{i}}$.
		\State ${v}_{prev}^{c_z}\leftarrow {v}_{g_i}$.
		\State ${v}_{new}^{c_z}\leftarrow {v}_{g_i}$.
		\State $\bm{w}_T^{g_i}\leftarrow \bm{w}_{last}^{g_{i}}$.
		\EndFor
		
		\EndFunction
	\end{algorithmic}
\end{algorithm}

\begin{algorithm}[htbp]
	\caption{Grouping Clients}
	\label{alg:groupclients}
	\begin{algorithmic}[1]
		\Require
		number of all clients $K$, learning rate ${\eta}$, local minibatch size $B$, number of local epochs $E$,global initial model $\bm{w}_{0}$, number of group $n$ pre-training hyper-parameter $\alpha$, weight of time $\beta$.
		\Ensure
		set of groups $G$			
		\Procedure {Grouping Clients}{}
		\State $\mathcal{M}\leftarrow$ Calculate $\mathcal{M}$ $ \verb|\\|$ref CALCULATE $\mathcal{M}$ 
		\State $[g_{1}.clients,...,g_{n}.clients]\leftarrow$~\textbf{SpectralClustering($\mathcal{M}$)}
		\State $G\leftarrow [g_{1},...,g_{n}]$
		\State \Return $G$
		\EndProcedure
	\end{algorithmic}
\end{algorithm}

\subsubsection{Communication Latency}
We consider such a communication scenario, where all the clients participating in the training are within the cell radius of the central server. 
Those who are out of range can not participate in training. 
There is path loss in the transmission link. 
Given an FDMA system with a total bandwidth of $\mathcal{W}$, for client $c_{i}$, its signal-to-noise ratio ($SNR$) is defined as follows~\cite{liangliang}:
\begin{equation}\label{fml:snr}
	{SNR}_i=\frac{P_{i} P_{L}^i}{N_0 \mathcal{W}}	
	,
\end{equation}
where $P_{i}$ is the transmission power (unit: $dbm$) from client $c_{i}$ to its group $g_{x}$, $N_0$ is the thermal noise variance (unit: $dbm/hz$), $P_{L}^i$ is the path loss (unit: $db$) between client $c_{i}$ and group $g_{x}$, we define $P_{L}^i$ as follows:
\begin{equation}\label{fml:ploss}
	P_{L}^i=100.7+23.5 \lg R	
	,
\end{equation}
Where $R$ is the distance (unit: $km$) between client $c_{i}$ and group $g_{x}$.

Based on~\eqref{fml:snr} and~\eqref{fml:ploss}, we define the transmission speed from client $c_{i}$ to group $g_{x}$ as follows:
\begin{equation}\label{fml:speed}
	C_i={\gamma}_i \mathcal{W} \log 2 (1+{SNR}_i)
	,
\end{equation}
where ${\gamma}_i \mathcal{W}$ represents the bandwidth allocated to client $c_{i}$. Based on~\eqref{fml:speed}, we define the communication latency of local update of client $c_{i}$ as:
\begin{equation}\label{fml:cm}
	t_{cm}^{i}=\frac{S_{model} }{C_i}	
	,
\end{equation}
where $S_{model}$ is the size of the model update. 
Because the transmission power of the physical facility where the group $g_{x}$ is located is relatively large, we ignore the communication latency from the group $g_{x}$ to the client $c_{i}$. 

\subsubsection{Model Update Latency}
We only cluster the classified clients once, so we take the mathematical expectation of each device's computation latency. 
Therefore, in the process of a model update, the model update latency of the client $c_{i}$ is defined by the following formula: 
\begin{equation}\label{fml:nul}
	t^{i}=t_{cm}^{i}+t_{excp}^i
	,
\end{equation}

Inspired by these CFL papers~\cite{cfl1}~\cite{cfl2}, we calculate the cosine similarity of the gradient update between clients to obtain their similarity. 
We define the cosine similarity between client $c_{i}$ and client $c_{j}$:
\begin{equation}\label{fml:cos}
	{Cosine}\left \langle {i,j}\right \rangle=\frac{\Delta\bm{w}^{i} \cdot \Delta\bm{w}^{j}}{\left \| \Delta\bm{w}^{i} \right \| \left \| \Delta\bm{w}^{j} \right \|}
	,
\end{equation}
where $\Delta\bm{w}^{i}$ is the vector of gradient update of client $c_{i}$, similarly, $\Delta\bm{w}^{j}$ is the vector of gradient update of client $c_{j}$. 

\begin{algorithm}[htbp]
	\caption{Similarity Matrix Calculation}
	\label{alg:simimatrix}
	\begin{algorithmic}[1]
		\Require
		(same as Algorithm 3)
		\Ensure
		similarity matrix $\mathcal{M}$			
		\Procedure {Calculate $\mathcal{M}$}{}
		\State $S_r\leftarrow$ set of all clients $R$
		\State ~\textbf{NormalizeTime($S_r$)}
		\For{each client $c_j$ in $S_{r}$ parallelly}
		\State $\Delta\bm{w}_0^{c_j}\leftarrow$~\textbf{PreTrainClient($c_j$,~$\bm{w}_0$)}		
		\EndFor
		\For{each client $c_j$ in $S_{r}$ parallelly}
		\State $\mathcal{L}\leftarrow$~\textbf{Similarity($c_j$,~$S_r$,~$\beta$)}
		\State $\mathcal{M}\leftarrow \mathcal{M} + [[\mathcal{L}]]$
		\EndFor
		\State \Return $\mathcal{M}$
		\EndProcedure
		\Statex
		\Function{Similarity}{$c$,~$S$,~$\beta$}
		\State $\mathcal{L}\leftarrow [\beta {t}^{c}]$.
		\For{each client $c_j$ in $S$ parallelly}
		\State $cosine\leftarrow $ ~\textsl{CosineSimilarity}($\Delta\bm{w}_0^{c}$,~$\Delta\bm{w}_0^{c_j}$)
		\State $\mathcal{L}\leftarrow \mathcal{L} + [cosine]$
		\EndFor
		\State \Return $\mathcal{L}$
		\EndFunction
		\Statex
		\Function{PreTrainClient}{$c$,~$\bm{w}$}
		\State $\hat{\bm{w}}\leftarrow\bm{w}$
		\State $\mathcal{B}\leftarrow$ split dataset of client $c$ into batches of size $B$.
		\For{each local epoch $e$ from $1$ to $E$}
		\For{each local batch $b \in \mathcal{B}$}
		\State $\bm{w}\leftarrow\bm{w}-\eta\nabla L(b;\bm{w})$.
		\EndFor
		\EndFor
		\State $\Delta\bm{w}\leftarrow\bm{w}-\hat{\bm{w}}$
		\State \Return ~\textbf{Flatten($\Delta\bm{w}$)}
		\EndFunction
		\Statex
		\Function{NormalizeTime}{$S$}
		\State $avg\leftarrow \sum_{c_j \in S} \frac{t_{cm}^{c_j}+t_{excp}^{c_j}}{\|S\|}$
		\State $var\leftarrow \sum_{c_j \in S} \frac{(t_{cm}^{c_j}+t_{excp}^{c_j}-avg)^2}{\|S\|}$
		\For{each client $c_j$ in $S$ parallelly}
		\State $t^{c_j}\leftarrow \frac{t_{c_j}^{cm}+t_{c_j}^{excp}-avg}{var}$
		\EndFor
		
		\EndFunction
		
	\end{algorithmic}
\end{algorithm}
\noindent$\displaystyle \textbf{Building Affinity Matrix}$

Before we build the affinity matrix, we should build the similarity matrix. In order not to make the latency with a small value lose its effect, we need to normalize these latencies. 
As shown in Algorithm \ref{alg:simimatrix}, we first calculate the variance and mean value of the model update latency of the clients to be classified, expressed by $var$ and $avg$ respectively, and then normalize the model update latency of the clients (line $3$, Line$24$ to line$28$): 
\begin{equation}\label{fml:nul}
	{t}_{n}^{i}=\frac{t^{i}-avg}{var}
	. 
\end{equation}
Finally, given a hyper-parameter weight $\beta$, $\beta {t}_{n}^{i}$ which represents the trade off between model update latency and gradient update on the same client $c_{i}$. 

We first leverage the~\eqref{fml:cos} to calculate the cosine similarities between the client $c_{i}$ and the clients to be classified. 
We use a vector $I$ ($I \subset \mathbb{R}^{1 \times v}$) to represent the cosine similarities of the client $c_{i}$. $v$ is the number of clients to be classified. Then we connect ${t}_{n}^{i}$ and vector $cosine$ as vector $L$ ($L \subset \mathbb{R}^{1 \times {(v+1)}}$). 
Finally, the vector $L$ is used as the client's data $c_{i}$ for classification. 
Similarly, the similarity matrix $M$ ($M \subset \mathbb{R}^{v \times {(v+1)}}$) is constructed by calculating the vector $L$ of all clients to be classified. At last, Gaussian similarity is used to construct an affinity matrix between clients based on the row vector of the similarity matrix $M$. The formula is as follows: 
\begin{equation}\label{fml:gos}
	{Gaussian}\left \langle {i,j}\right \rangle=e^{-\frac{\left \| L^i - L^j \right \|}{2 {\sigma}^2}}
	,
\end{equation}
 
In this paper, we use the average aggregation strategy of \textit{FedAvg}.

To verify our CSAFL framework is reasonable, we will evaluate it based on several real federated datasets in the next section.

\section{evaluation}\label{sec:evaluation}
In this section, we introduce the experimental results of the CSAFL framework. 
We evaluate our experiments based on four open federated datasets. 
We show the details of our experiments in Section~\ref{subsec:A}. Then, in Section~\ref{subsec:B}, we present the performance improvement of our method compared to the baselines. 
In order to further verify the effectiveness of our grouping strategy and update strategy, specifically, we design comparative experiments, which are introduced in Section~\ref{subsec:C} and Section~\ref{subsec:D}, respectively. 
For comparison, we fix the hyper-parameter time budget $T$ in all experiments. 
Specifically, each communication round in the training process of each experiment has the same time budget. 
Our code is based on TensorFlow~\cite{tensorflow}.

\begin{figure*}[h]
	\centering
	\includegraphics[width=18cm]{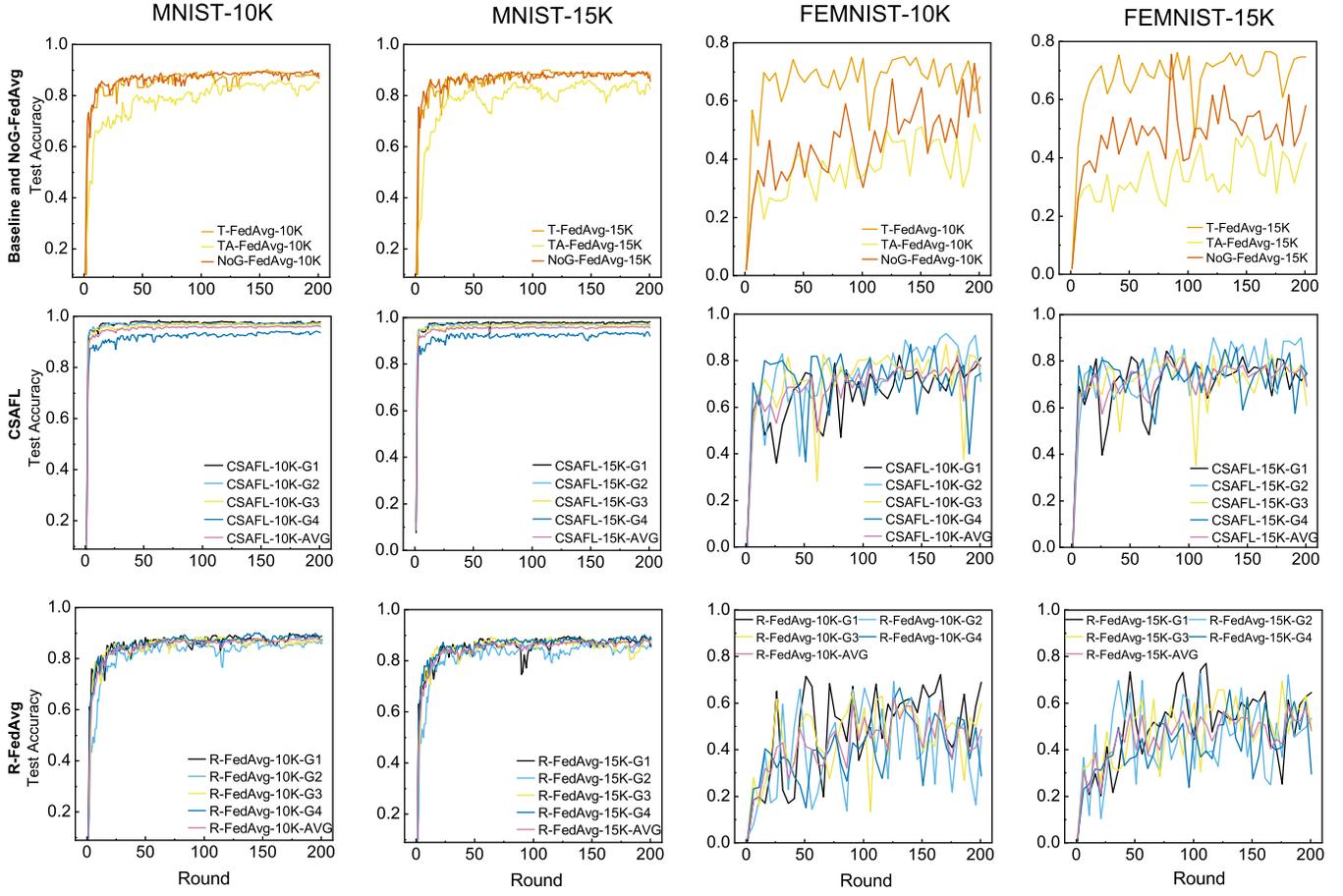}
	\vspace{-3mm}
	\caption{The accuracy curves of CSAFL, \textit{R-FedAvg}, baselines and \textit{NoG-FedAvg} on MNIST and FEMNIST. 10K=10000, 15K=15000, unit:ms.}
	\label{fig:straggler}
\end{figure*}
\begin{figure*}[t]
	\centering
	\vspace{-3mm}
	\includegraphics[width=18cm]{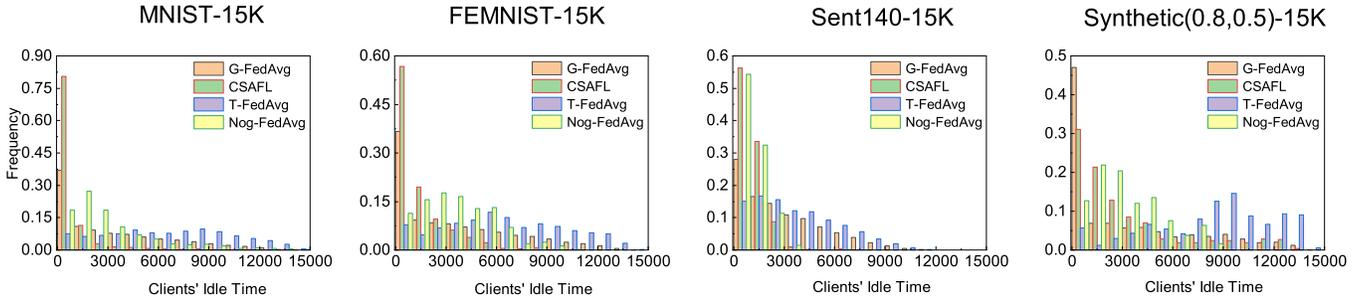}
	\vspace{-3mm}
	\caption{The frequency distribution of the clients' idle time. 10K=10000, 15K=15000, unit: ms.}
	\label{fig:eval1}
\end{figure*}
\subsection[A]{Experimental Setup}\label{subsec:A}
The parameter of FL setting we used in this paper is described in~\ref{sec:motivation}, and except these, we set model delay threshold $L = 4$, We evaluated our experiments on four federal datasets, where including two image classification datasets and a synthetic dataset, an emotion analysis dataset. 
We use a convex multinomial logistic regression (MCLR) model for training the first three datasets, and we use a Long Short-Term Memory (LSTM) model for training the last emotion analysis task. 
The statistical information of datasets and models is shown in Table~\ref{table1}. The details are as follows:

\begin{table*}[t]
	\centering  
	\caption{CSAFL and its comparison algorithms are evaluated on four federated datasets, and this table shows the accuracy of the model with these algorithms}  
	\label{table2}  
	\begin{tabular}{|l|cc|ccc|cc|}
		\hline
		Dataset-Time Budget & T-FedAvg      & TA-FedAvg     & G-FedAvg & GA-FedAvg & R-FedAvg & NoG-FedAvg & CSAFL \\ \hline
		MNIST-10000         & \textbf{90.0} & 86.1          & 96.3     & 95.8      & 88.0     & 89.9       & 96.2  \\
		MNIST-15000         & \textbf{90.0} & 86.0          & 96.2     & 95.6      & 88.3     & 89.6       & 96.2  \\
		FEMNIST-10000       & \textbf{77.1} & 55.1          & 88.6     & 77.1      & 62.4     & 74.1       & 84.2  \\
		FEMNIST-15000       & \textbf{77.4} & 51.1          & 89.7     & 77.7      & 63.0     & 77.2       & 85.5  \\
		Synthetic-10000     & 26.6          & 33.0          & 88.1     & 64.8      & 31.9     & 40.5       & 66.2  \\
		Synthetic-15000     & 20.4          & \textbf{53.6} & 71.4     & 62.3      & 35.6     & 52.9       & 64.1  \\
		Sent140-10000       & \textbf{71.3} & 62.3          & 71.0     & 63.6      & 66.4     & 69.5       & 68.0  \\
		Sent140-15000       & \textbf{71.1} & 62.5          & 69.7     & 62.8      & 65.3     & 70.1       & 69.0  \\ \hline
	\end{tabular}
\end{table*}
\noindent$\displaystyle \textbf{Datasets and Models.}$ Our experiments are based on four non-IID datasets which are class-imbalanced and data-imbalanced, and we leverage the appropriate models to train these datasets, the details follow as: 
\begin{itemize}
	\item MNIST~\cite{lecun2010mnist}: a handwritten digits data set, which is divided into $0$ to $9$ the $10$ categories of numbers, each data is a flatted $784$ dimensional ($28$x$28$) pixel image. According to the power law, we assign data to $1000$ clients, and each client has two types of digits. We use the MCLR model, and the input of the model is pixel images, and the output is $10$ labels of digitals. 
	\item FEMNIST: a $62$ class federated extended MNIST dataset, which is based on EMNIST~\cite{cohen2017emnist} dataset construction, and only sampling $a$ to $j$ these $10$ types of lowercase characters. We divide data into $200$ clients by the power law, and each client contains five types. We also use the MCLR model, and the input and output of the model are similar to MNIST. 
	\item Synthetic: Shamir \textit{et al.} propose a synthetic federated dataset~\cite{syntheticdata}. We set (a,b)=($0.8,0.5$) and divide data into $100$ clients. Similarly, for this synthetic data set, we use the MCLR model to test it. 
	\item Sentiment140 (Sent140)~\cite{sent140}: a data set based on Twitter users’ emoticons to express the sentiment, each user as a client, we use a two-layer logistic regression model with $256$ hidden units to train emoticons based on $100$ Twitter users.
\end{itemize}

\noindent$\displaystyle \textbf{Baseline.}$ We have two baseline methods:
\begin{itemize}
	\item a synchronous method \textit{FedAvg} with the time budget that we call it \textit{T-FedAvg}.
	\item The asynchronous method is based on T-fedAvg, which we call it \textit{TA-FedAvg}.
\end{itemize}

\noindent$\displaystyle \textbf{Comparison methods.}$ In order to Significantly illustrate the performance of our method, we designed six comparison experiments. The specific FL settings are similar to our method. We will elaborate in the following sections.

\noindent$\displaystyle \textbf{Metrics.}$ We have two metrics:

\begin{enumerate}
	\item Frequency of clients’ idle time: We record each client's idle time in the training process of the synchronous round. In order to evaluate the effectiveness of the CSAFL framework in mitigating the straggler effect, we evaluate the frequency of clients who wait for more than $60\%$ of the time budget.
	\item Accuracy: CSAFL framework is based on the clustered algorithm, and its extended comparative experiments are also divided into groups for testing. The accuracy of each client is calculated based on the group model of the group to which the client belongs, so we use the weighted test accuracy to evaluate each group model for intuitively expressing the overall performance of grouped experiments. 
\end{enumerate}

\subsection[B]{Experimental Results}\label{subsec:B}
As shown in Table~\ref{table2}, we present the experimental results of our framework with the baseline methods. The results show that our method is superior to the baseline method in  MNIST, FEMNIST, Synthetic. Especially, CSAFL improves test accuracy by $43.7\%$ on synthetic( $0.8,0.5$) with $L = 15000$. besides, the accuracy of \textit{TA-FedAvg} is better than that of \textit{T-FedAvg}, which indicates the influence of the straggler effect on synchronous FL. Although the average test accuracy of CSAFL is not better than that of \textit{T-FedAvg}g on Sent140, we can observe from Fig.~\ref{fig:straggler} that the group curve of CSAFL is superior. 

\begin{figure*}[h]
	\centering
	\vspace{-3mm}
	\includegraphics[width=18cm]{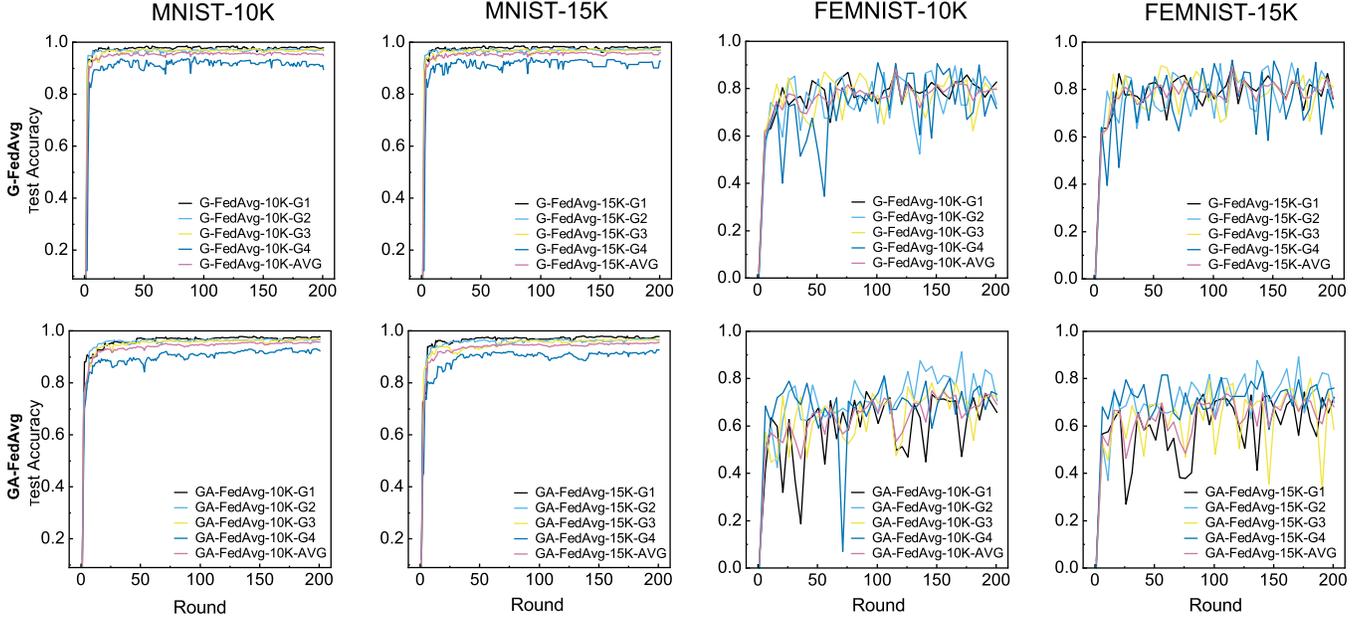}
	\vspace{-3mm}
	\caption{The accuracy curves of \textit{G-FedAvg}, \textit{GA-FedAvg} on MNIST and FEMNIST. 10K=10000, 15K=15000, unit: ms.}

	\label{fig:eval2}
\end{figure*}

More detail that each communication round's test accuracy is shown in Fig.~\ref{fig:straggler} and Fig.~\ref{fig:eval2}. We observe that our method is roughly as fast as synchronous FL and faster than asynchronous FL in the convergence speed of the model on FEMNIST. That is, our CSAFL framework can converge infinite communication rounds. On the other hand, the convergence rate of CSAFL is about the same as that of the \textit{T-FedAvg} and \textit{TA-FedAvg} on MNIST, which may be that different model update strategies will not cause significant divergence of gradients on MNIST. 

\subsection[C]{Effects of The Grouping Strategy}\label{subsec:C}
To demonstrate the effectiveness of the grouping strategy, we set up two groups of comparative experiments, as follows:
\begin{itemize}
	\item Under the same update strategy, we use the random grouping method R-FedAvg based on T-FedAvg to compare with CSAFL. As shown in Table~\ref{table2}, the test accuracy of CSAFL is higher than that of the R-FedAvg at least $7.9\%$ on MNIST, FEMNIST, Synthetic. As for Sent140, We can observe that in Fig.~\ref{fig:straggler}, the curve of CSAFL is higher than R-FedAvg in overall accuracy. In short,  our grouping strategy is effective.
	\item Similarly, based on the same update strategy, we design a \textit{NoG-FedAvg} without grouping. Compared with CSAFL, the results show that the curve fluctuation of CSAFL is smaller than that of \textit{NoG-FedAvg} in Fig.~\ref{fig:straggler} and Fig.~\ref{fig:eval2}. 
\end{itemize}

\subsection[D]{Effects of The Update Strategy}\label{subsec:D}
In order to verify the effectiveness of the update strategy, we designed two groups of comparative experiments:
\begin{itemize}
	\item For different update strategies in the case of grouping, we design the synchronous FL and asynchronous FL based on FedAvg, which are named \textit{G-FedAvg} and \textit{GA-FedAvg}, respectively. As shown in Table~\ref{table2}, we can observe that our method's experimental results are better than other update strategies.
	\item Without grouping, we use \textit{GA-FedAvg} and \textit{T-FedAvg} to compare with \textit{TA-FedAvg}, and the results show the effectiveness of our strategy
\end{itemize}

\subsection[E]{effects of the update strategy}\label{subsec:E}
In order to verify that CSAFL can mitigate the straggler effect, as shown in Fig.~\ref{fig:eval1}, in the time budget $L = 15000ms$, we can observe the frequency distribution of clients’ idle time. The results show that CSAFL can significantly mitigate the straggler effect. Especially on the MNIST, FEMNIST, Sent140, these frequencies of the clients’ idle time is close to zero after $6000 ms$. On the Synthetic, we can observe that the frequencies of T-FedAvg's clients’ idle time are greater than $0.07$ after $9000 ms$, which also shows that the conclusion in Section D is correct. That is, the straggler effect affects the accuracy of the synchronization updates.

\section{conclusion}\label{sec:conclusion}
In this work, we propose a new clustered semi-asynchronous federated learning framework (CSAFL), which can effectively mitigate the straggler effect and improve the accuracy of the asynchronous FL. Based on four datasets, our evaluation experiments show that the CSAFL framework is better than asynchronous FL on accuracy, convergence speed, and CSAFL is an effective solution to the straggler effect in synchronous FL. We further prove the effectiveness of the grouping strategy and update mechanism. 

In the future, we will explore the following directions:
\begin{enumerate}
	\item The advantages of our framework in privacy protection.
	\item We will further expand the trade-off of time and direction; personalized learning beta is based on various data sets.
	\item We will explore a better strategy to aggregate global models.
\end{enumerate}

\bibliographystyle{IEEEtran}
\balance
\bibliography{IEEEabrv,references}

\begin{thebibliography}{10}
\providecommand{\url}[1]{#1}
\csname url@samestyle\endcsname
\providecommand{\newblock}{\relax}
\providecommand{\bibinfo}[2]{#2}
\providecommand{\BIBentrySTDinterwordspacing}{\spaceskip=0pt\relax}
\providecommand{\BIBentryALTinterwordstretchfactor}{4}
\providecommand{\BIBentryALTinterwordspacing}{\spaceskip=\fontdimen2\font plus
\BIBentryALTinterwordstretchfactor\fontdimen3\font minus
  \fontdimen4\font\relax}
\providecommand{\BIBforeignlanguage}[2]{{%
\expandafter\ifx\csname l@#1\endcsname\relax
\typeout{** WARNING: IEEEtran.bst: No hyphenation pattern has been}%
\typeout{** loaded for the language `#1'. Using the pattern for}%
\typeout{** the default language instead.}%
\else
\language=\csname l@#1\endcsname
\fi
#2}}
\providecommand{\BIBdecl}{\relax}
\BIBdecl

\bibitem{mcmahan2017communication}
B.~McMahan, E.~Moore, D.~Ramage, S.~Hampson, and B.~A. y~Arcas,
  ``Communication-efficient learning of deep networks from decentralized
  data,'' in \emph{Proceedings of the 20th International Conference on
  Artificial Intelligence and Statistics (AISTATS)}, 2017, pp. 1273--1282.

\bibitem{zhao2018federated}
Y.~Zhao, M.~Li, L.~Lai, N.~Suda, D.~Civin, and V.~Chandra, ``Federated learning
  with non-iid data,'' \emph{arXiv preprint arXiv:1806.00582}, 2018.

\bibitem{li2020federated}
T.~Li, A.~K. Sahu, A.~Talwalkar, and V.~Smith, ``Federated learning:
  Challenges, methods, and future directions,'' \emph{IEEE Signal Processing
  Magazine}, vol.~37, no.~3, pp. 50--60, 2020.

\bibitem{privacy2019}
M.~Hao, H.~Li, G.~Xu, S.~Liu, and H.~Yang, ``Towards efficient and
  privacy-preserving federated deep learning,'' in \emph{2019 {IEEE}
  International Conference on Communications, {ICC} 2019, Shanghai, China, May
  20-24, 2019}.\hskip 1em plus 0.5em minus 0.4em\relax {IEEE}, 2019, pp. 1--6.

\bibitem{privacy2020}
C.~Fang, Y.~Guo, N.~Wang, and A.~Ju, ``Highly efficient federated learning with
  strong privacy preservation in cloud computing,'' \emph{Comput. Secur.},
  vol.~96, p. 101889, 2020.

\bibitem{privacy2021}
V.~Mothukuri, R.~M. Parizi, S.~Pouriyeh, Y.~Huang, A.~Dehghantanha, and
  G.~Srivastava, ``A survey on security and privacy of federated learning,''
  \emph{Future Gener. Comput. Syst.}, vol. 115, pp. 619--640, 2021.

\bibitem{privacyapp}
J.~Feng, C.~Rong, F.~Sun, D.~Guo, and Y.~Li, ``{PMF:} {A} privacy-preserving
  human mobility prediction framework via federated learning,'' \emph{Proc.
  {ACM} Interact. Mob. Wearable Ubiquitous Technol.}, vol.~4, no.~1, pp.
  10:1--10:21, 2020.

\bibitem{stragglereffect}
T.~T. Vu, D.~T. Ngo, H.~Q. Ngo, M.~N. Dao, N.~H. Tran, and R.~H. Middleton,
  ``User selection approaches to mitigate the straggler effect for federated
  learning on cell-free massive {MIMO} networks,'' \emph{CoRR}, vol.
  abs/2009.02031, 2020.

\bibitem{device}
W.~Shi, S.~Zhou, and Z.~Niu, ``Device scheduling with fast convergence for
  wireless federated learning,'' in \emph{2020 {IEEE} International Conference
  on Communications, {ICC} 2020, Dublin, Ireland, June 7-11, 2020}.\hskip 1em
  plus 0.5em minus 0.4em\relax {IEEE}, 2020, pp. 1--6.

\bibitem{Li2020On}
X.~Li, K.~Huang, W.~Yang, S.~Wang, and Z.~Zhang, ``On the convergence of fedavg
  on non-iid data,'' in \emph{Proceedings of the 8th International Conference
  on Learning Representations (ICLR)}, 2020.

\bibitem{duan2019astraea}
M.~Duan, D.~Liu, X.~Chen, Y.~Tan, J.~Ren, L.~Qiao, and L.~Liang, ``Astraea:
  Self-balancing federated learning for improving classification accuracy of
  mobile deep learning applications,'' in \emph{Proceedings of the IEEE 37th
  International Conference on Computer Design (ICCD)}.\hskip 1em plus 0.5em
  minus 0.4em\relax IEEE, 2019, pp. 246--254.

\bibitem{xie2019fedasyn}
\BIBentryALTinterwordspacing
I.~G. Cong~Xie, Sanmi~Koyejo, ``Asynchronous federated optimization,''
  \emph{CoRR}, vol. abs/1903.03934, 2019. [Online]. Available:
  \url{http://arxiv.org/abs/1903.03934}
\BIBentrySTDinterwordspacing

\bibitem{aso-fed}
Y.~Chen, Y.~Ning, and H.~Rangwala, ``Asynchronous online federated learning for
  edge devices,'' \emph{CoRR}, vol. abs/1911.02134, 2019.

\bibitem{spectral}
P.~K. Chan, M.~D.~F. Schlag, and J.~Y. Zien, ``Spectral k-way ratio-cut
  partitioning and clustering,'' \emph{{IEEE} Trans. Comput. Aided Des. Integr.
  Circuits Syst.}, vol.~13, no.~9, pp. 1088--1096, 1994.

\bibitem{noconsider2}
J.~Ren, G.~Yu, and G.~Ding, ``Accelerating {DNN} training in wireless federated
  edge learning systems,'' \emph{{IEEE} J. Sel. Areas Commun.}, vol.~39, no.~1,
  pp. 219--232, 2021.

\bibitem{noconsider3}
T.~Sery, N.~Shlezinger, K.~Cohen, and Y.~C. Eldar, ``{COTAF:} convergent
  over-the-air federated learning,'' in \emph{{IEEE} Global Communications
  Conference, {GLOBECOM} 2020, Virtual Event, Taiwan, December 7-11,
  2020}.\hskip 1em plus 0.5em minus 0.4em\relax {IEEE}, 2020, pp. 1--6.

\bibitem{li2018federated}
T.~Li, A.~K. Sahu, M.~Zaheer, M.~Sanjabi, A.~Talwalkar, and V.~Smith,
  ``Federated optimization in heterogeneous networks,'' in \emph{Proceedings of
  the 3rd SysML Conference}, 2020.

\bibitem{lifedavgnoniid}
X.~Li, K.~Huang, W.~Yang, S.~Wang, and Z.~Zhang, ``On the convergence of fedavg
  on non-iid data,'' in \emph{8th International Conference on Learning
  Representations, {ICLR} 2020, Addis Ababa, Ethiopia, April 26-30,
  2020}.\hskip 1em plus 0.5em minus 0.4em\relax OpenReview.net, 2020.

\bibitem{codefl}
S.~Dhakal, S.~Prakash, Y.~Yona, S.~Talwar, and N.~Himayat, ``Coded federated
  learning,'' \emph{CoRR}, vol. abs/2002.09574, 2020.

\bibitem{safafed}
W.~Wu, L.~He, W.~Lin, R.~Mao, and S.~A. Jarvis, ``{SAFA:} a semi-asynchronous
  protocol for fast federated learning with low overhead,'' \emph{CoRR}, vol.
  abs/1910.01355, 2019.

\bibitem{fedasyn20201}
Y.~Chen, X.~Sun, and Y.~Jin, ``Communication-efficient federated deep learning
  with layerwise asynchronous model update and temporally weighted
  aggregation,'' \emph{{IEEE} Trans. Neural Networks Learn. Syst.}, vol.~31,
  no.~10, pp. 4229--4238, 2020.

\bibitem{fedasyn2020}
X.~Lu, Y.~Liao, P.~Li{\`{o}}, and P.~Hui, ``Privacy-preserving asynchronous
  federated learning mechanism for edge network computing,'' \emph{{IEEE}
  Access}, vol.~8, pp. 48\,970--48\,981, 2020.

\bibitem{caldas2018leaf}
S.~Caldas, P.~Wu, T.~Li, J.~Kone{\v{c}}n{\`y}, H.~B. McMahan, V.~Smith, and
  A.~Talwalkar, ``Leaf: A benchmark for federated settings,'' \emph{arXiv
  preprint arXiv:1812.01097}, 2018.

\bibitem{Avishek2020cluster}
A.~Ghosh, J.~Chung, D.~Yin, and K.~Ramchandran, ``An efficient framework for
  clustered federated learning,'' in \emph{Advances in Neural Information
  Processing Systems 33: Annual Conference on Neural Information Processing
  Systems 2020, NeurIPS 2020, December 6-12, 2020, virtual}, H.~Larochelle,
  M.~Ranzato, R.~Hadsell, M.~Balcan, and H.~Lin, Eds., 2020.

\bibitem{kmeans}
D.~Pollard, ``Quantization and the method of k -means,'' \emph{{IEEE} Trans.
  Inf. Theory}, vol.~28, no.~2, pp. 199--204, 1982.

\bibitem{cfl1}
C.~Briggs, Z.~Fan, and P.~Andras, ``Federated learning with hierarchical
  clustering of local updates to improve training on non-iid data,'' in
  \emph{2020 International Joint Conference on Neural Networks, {IJCNN} 2020,
  Glasgow, United Kingdom, July 19-24, 2020}.\hskip 1em plus 0.5em minus
  0.4em\relax {IEEE}, 2020, pp. 1--9.

\bibitem{cfl2}
F.~Sattler, K.~M{\"{u}}ller, T.~Wiegand, and W.~Samek, ``On the byzantine
  robustness of clustered federated learning,'' in \emph{2020 {IEEE}
  International Conference on Acoustics, Speech and Signal Processing, {ICASSP}
  2020, Barcelona, Spain, May 4-8, 2020}.\hskip 1em plus 0.5em minus
  0.4em\relax {IEEE}, 2020, pp. 8861--8865.

\bibitem{shift}
K.~Lee, M.~Lam, R.~Pedarsani, D.~S. Papailiopoulos, and K.~Ramchandran,
  ``Speeding up distributed machine learning using codes,'' \emph{{IEEE} Trans.
  Inf. Theory}, vol.~64, no.~3, pp. 1514--1529, 2018.

\bibitem{liangliang}
L.~{Liang}, G.~{Feng}, and Y.~{Zhang}, ``Integrated interference coordination
  for relay-aided cellular ofdma system,'' in \emph{2011 IEEE International
  Conference on Communications (ICC)}, 2011.

\bibitem{tensorflow}
M.~Abadi, P.~Barham, J.~Chen, Z.~Chen, A.~Davis, J.~Dean, and M.~Devin,
  ``Tensorflow: {A} system for large-scale machine learning,'' in \emph{12th
  {USENIX} Symposium on Operating Systems Design and Implementation, {OSDI}
  2016, Savannah, GA, USA, November 2-4, 2016}.\hskip 1em plus 0.5em minus
  0.4em\relax {USENIX} Association, 2016, pp. 265--283.

\bibitem{lecun2010mnist}
Y.~LeCun, C.~Cortes, and C.~Burges, ``Mnist handwritten digit database,''
  \emph{ATT Labs [Online]. Available: http://yann. lecun. com/exdb/mnist},
  vol.~2, 2010.

\bibitem{cohen2017emnist}
G.~Cohen, S.~Afshar, J.~Tapson, and A.~van Schaik, ``Emnist: Extending mnist to
  handwritten letters,'' in \emph{Proceedings of the 2017 International Joint
  Conference on Neural Networks (IJCNN)}.\hskip 1em plus 0.5em minus
  0.4em\relax IEEE, 2017, pp. 2921--2926.

\bibitem{syntheticdata}
O.~Shamir, N.~Srebro, and T.~Zhang, ``Communication-efficient distributed
  optimization using an approximate newton-type method,'' in \emph{Proceedings
  of the 31th International Conference on Machine Learning, {ICML} 2014,
  Beijing, China, 21-26 June 2014}, ser. {JMLR} Workshop and Conference
  Proceedings, vol.~32.\hskip 1em plus 0.5em minus 0.4em\relax JMLR.org, 2014,
  pp. 1000--1008.

\bibitem{sent140}
T.~Sahni, C.~Chandak, N.~R. Chedeti, and M.~Singh, ``Efficient twitter
  sentiment classification using subjective distant supervision,'' in \emph{9th
  International Conference on Communication Systems and Networks, {COMSNETS}
  2017, Bengaluru, India, January 4-8, 2017}.\hskip 1em plus 0.5em minus
  0.4em\relax {IEEE}, 2017, pp. 548--553.

\end{thebibliography}


\begin{thebibliography}{10}
\providecommand{\url}[1]{#1}
\csname url@samestyle\endcsname
\providecommand{\newblock}{\relax}
\providecommand{\bibinfo}[2]{#2}
\providecommand{\BIBentrySTDinterwordspacing}{\spaceskip=0pt\relax}
\providecommand{\BIBentryALTinterwordstretchfactor}{4}
\providecommand{\BIBentryALTinterwordspacing}{\spaceskip=\fontdimen2\font plus
\BIBentryALTinterwordstretchfactor\fontdimen3\font minus
  \fontdimen4\font\relax}
\providecommand{\BIBforeignlanguage}[2]{{%
\expandafter\ifx\csname l@#1\endcsname\relax
\typeout{** WARNING: IEEEtran.bst: No hyphenation pattern has been}%
\typeout{** loaded for the language `#1'. Using the pattern for}%
\typeout{** the default language instead.}%
\else
\language=\csname l@#1\endcsname
\fi
#2}}
\providecommand{\BIBdecl}{\relax}
\BIBdecl

\bibitem{konevcny2016federated}
J.~Kone{\v{c}}n{\`y}, H.~B. McMahan, F.~X. Yu, P.~Richt{\'a}rik, A.~T. Suresh,
  and D.~Bacon, ``Federated learning: Strategies for improving communication
  efficiency,'' \emph{arXiv preprint arXiv:1610.05492}, 2016.

\bibitem{mcmahan2017communication}
B.~McMahan, E.~Moore, D.~Ramage, S.~Hampson, and B.~A. y~Arcas,
  ``Communication-efficient learning of deep networks from decentralized
  data,'' in \emph{Proceedings of the 20th International Conference on
  Artificial Intelligence and Statistics (AISTATS)}, 2017, pp. 1273--1282.

\bibitem{bonawitz2019towards}
K.~Bonawitz, H.~Eichner, W.~Grieskamp, D.~Huba, A.~Ingerman, V.~Ivanov,
  C.~Kiddon, J.~Konecny, S.~Mazzocchi, H.~B. McMahan \emph{et~al.}, ``Towards
  federated learning at scale: System design,'' in \emph{Proceedings of the 2nd
  SysML Conference}, 2019.

\bibitem{yang2019federated}
Q.~Yang, Y.~Liu, T.~Chen, and Y.~Tong, ``Federated machine learning: Concept
  and applications,'' \emph{ACM Transactions on Intelligent Systems and
  Technology (TIST)}, vol.~10, no.~2, pp. 1--19, 2019.

\bibitem{li2020federated}
T.~Li, A.~K. Sahu, A.~Talwalkar, and V.~Smith, ``Federated learning:
  Challenges, methods, and future directions,'' \emph{IEEE Signal Processing
  Magazine}, vol.~37, no.~3, pp. 50--60, 2020.

\bibitem{li2014scaling}
M.~Li, D.~G. Andersen, J.~W. Park, A.~J. Smola, A.~Ahmed, V.~Josifovski,
  J.~Long, E.~J. Shekita, and B.-Y. Su, ``Scaling distributed machine learning
  with the parameter server,'' in \emph{Proceedings of the 11th USENIX
  Symposium on Operating Systems Design and Implementation (OSDI)}, 2014, pp.
  583--598.

\bibitem{zhao2018federated}
Y.~Zhao, M.~Li, L.~Lai, N.~Suda, D.~Civin, and V.~Chandra, ``Federated learning
  with non-iid data,'' \emph{arXiv preprint arXiv:1806.00582}, 2018.

\bibitem{krizhevsky2009learning}
A.~Krizhevsky and G.~Hinton, ``Learning multiple layers of features from tiny
  images,'' Citeseer, Tech. Rep., 2009.

\bibitem{Simonyan15}
K.~Simonyan and A.~Zisserman, ``Very deep convolutional networks for
  large-scale image recognition,'' in \emph{Proceedings of the 3rd
  International Conference on Learning Representations (ICLR)}.\hskip 1em plus
  0.5em minus 0.4em\relax IEEE, 2015.

\bibitem{sattler2019robust}
F.~Sattler, S.~Wiedemann, K.-R. M{\"u}ller, and W.~Samek, ``Robust and
  communication-efficient federated learning from {non-i.i.d.} data,''
  \emph{IEEE Transactions on Neural Networks and Learning Systems (TNNLS)}, pp.
  1--14, 2019.

\bibitem{Li2020On}
X.~Li, K.~Huang, W.~Yang, S.~Wang, and Z.~Zhang, ``On the convergence of fedavg
  on non-iid data,'' in \emph{Proceedings of the 8th International Conference
  on Learning Representations (ICLR)}, 2020.

\bibitem{duan2019astraea}
M.~Duan, D.~Liu, X.~Chen, Y.~Tan, J.~Ren, L.~Qiao, and L.~Liang, ``Astraea:
  Self-balancing federated learning for improving classification accuracy of
  mobile deep learning applications,'' in \emph{Proceedings of the IEEE 37th
  International Conference on Computer Design (ICCD)}.\hskip 1em plus 0.5em
  minus 0.4em\relax IEEE, 2019, pp. 246--254.

\bibitem{sattler2020clustered}
F.~Sattler, K.-R. M{\"u}ller, and W.~Samek, ``Clustered federated learning:
  Model-agnostic distributed multitask optimization under privacy
  constraints,'' \emph{IEEE Transactions on Neural Networks and Learning
  Systems (TNNLS)}, pp. 1--13, 2020.

\bibitem{li2018federated}
T.~Li, A.~K. Sahu, M.~Zaheer, M.~Sanjabi, A.~Talwalkar, and V.~Smith,
  ``Federated optimization in heterogeneous networks,'' in \emph{Proceedings of
  the 3rd SysML Conference}, 2020.

\bibitem{lecun1998gradient}
Y.~LeCun, L.~Bottou, Y.~Bengio, and P.~Haffner, ``Gradient-based learning
  applied to document recognition,'' \emph{Proceedings of the IEEE}, vol.~86,
  no.~11, pp. 2278--2324, 1998.

\bibitem{dwork2015reusable}
C.~Dwork, V.~Feldman, M.~Hardt, T.~Pitassi, O.~Reingold, and A.~Roth, ``The
  reusable holdout: Preserving validity in adaptive data analysis,''
  \emph{Science}, vol. 349, no. 6248, pp. 636--638, 2015.

\end{thebibliography}

\end{document}